\PassOptionsToPackage{dvipsnames}{xcolor}
\documentclass[singlecolumn]{NTUAI4XPaperEN}

\usepackage{mathtools}
\usepackage{placeins}
\usepackage[english]{babel}
\usepackage{xspace}
\setthemecolor{19365b}

\usepackage{booktabs}     % Professional tables
\usepackage{algorithm}
\usepackage{algpseudocode}

\usepackage{microtype}      % microtypography

\usepackage{graphicx}
\usepackage{subcaption} % for subfigures

\usepackage{amsthm}

\usepackage{multirow}
\usepackage{makecell}
\usepackage{titletoc}

\crefname{section}{\S}{\S}
\crefname{table}{Tab.}{Tabs.}
\crefname{figure}{Fig.}{Figs.}

\titlecontents{section}
[2em]
{\bfseries}
{\contentslabel{1.8em}}
{}
{\titlerule*[0.8pc]{.}\contentspage}

\titlecontents{subsection}
[4.5em]
{}
{\contentslabel{3em}}
{}
{\titlerule*[0.8pc]{.}\contentspage}

\definecolor{Light}{rgb}{0.92, 0.99, 0.95}
\definecolor{mygray}{HTML}{E6E6FA}
\newcommand{\method}{\textsc{CuRB}\xspace}

\newcommand{\rdg}{\mbox{RDG}\xspace}
\newcommand{\crdg}{\mbox{cRDG}\xspace}
\newcommand{\crdgs}{\mbox{cRDG}}

\definecolor{gvnavy}{RGB}{1,33,105}
\definecolor{gvamber}{RGB}{232,163,61}
\definecolor{gvamberd}{RGB}{199,127,26}
\definecolor{gvice}{RGB}{242,245,250}
\definecolor{gvgreen}{RGB}{63,128,63}
\definecolor{gvred}{RGB}{178,34,34}
\definecolor{gvgray}{RGB}{107,118,137}
\definecolor{gvblue}{RGB}{53,108,170}
\definecolor{rowhl}{RGB}{251,241,221}
\definecolor{rowref}{RGB}{238,238,238}

\setthemecolor{1a2a6c}

\titlelogocontent{%
	\begin{minipage}{\textwidth}
		\begin{minipage}[c]{0.35\textwidth}
			\raggedright
			\includegraphics[height=1cm]{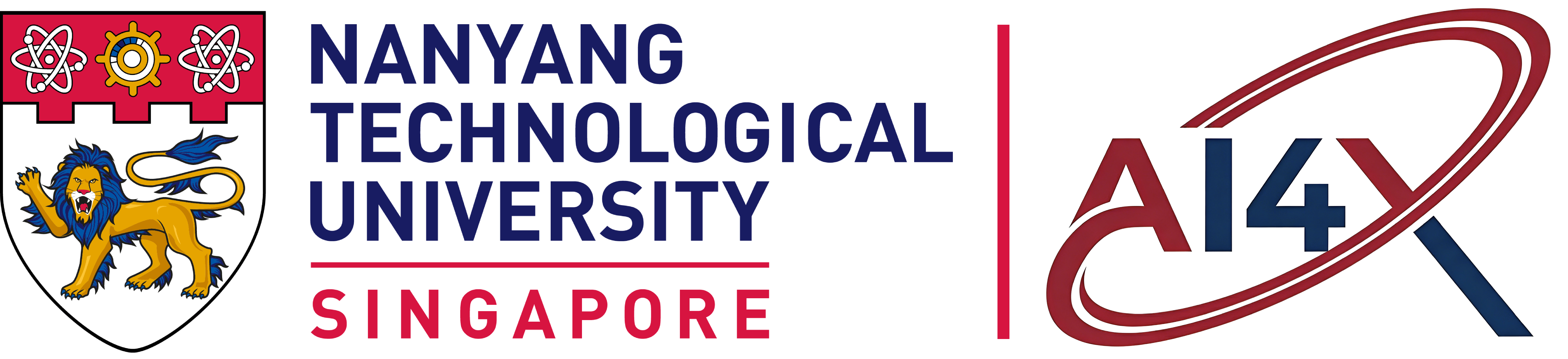}% 团队logo
		\end{minipage}%
		\hfill
		\begin{minipage}[c]{0.60\textwidth}
			\raggedleft
			\includegraphics[height=1cm]{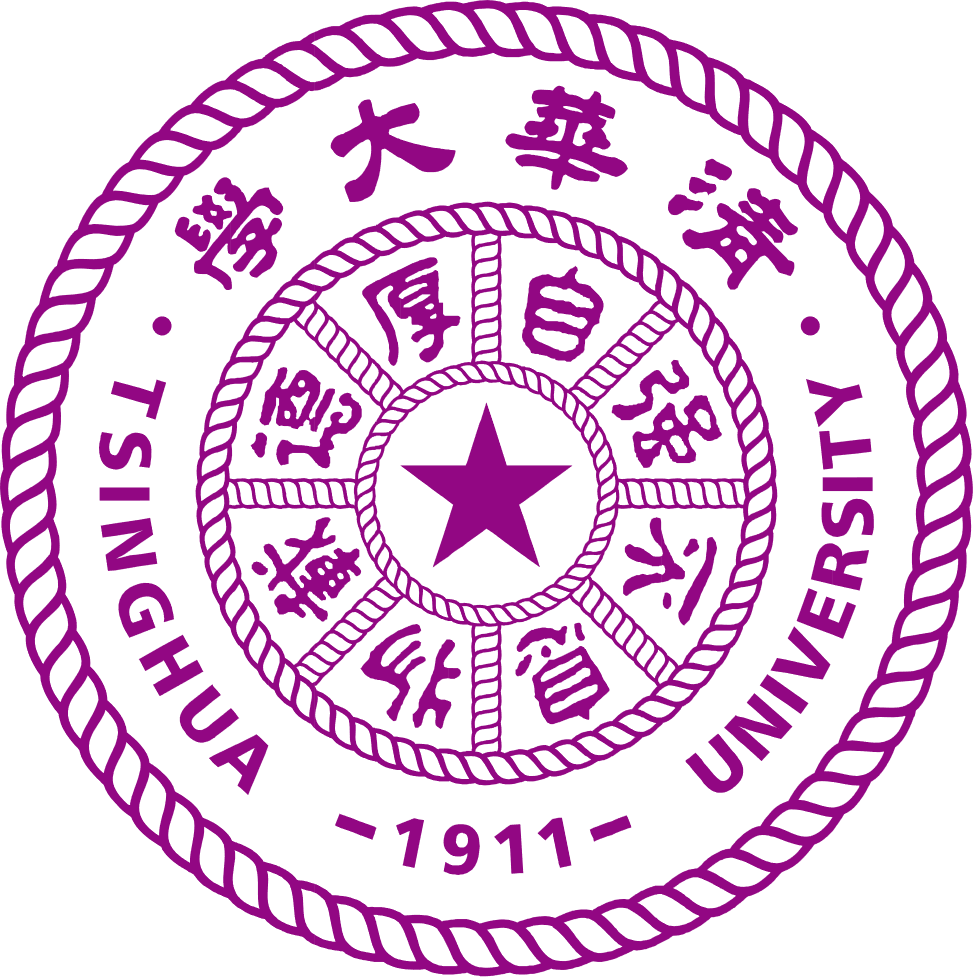}\hspace{0.6em}% 学校1
			\includegraphics[height=1cm]{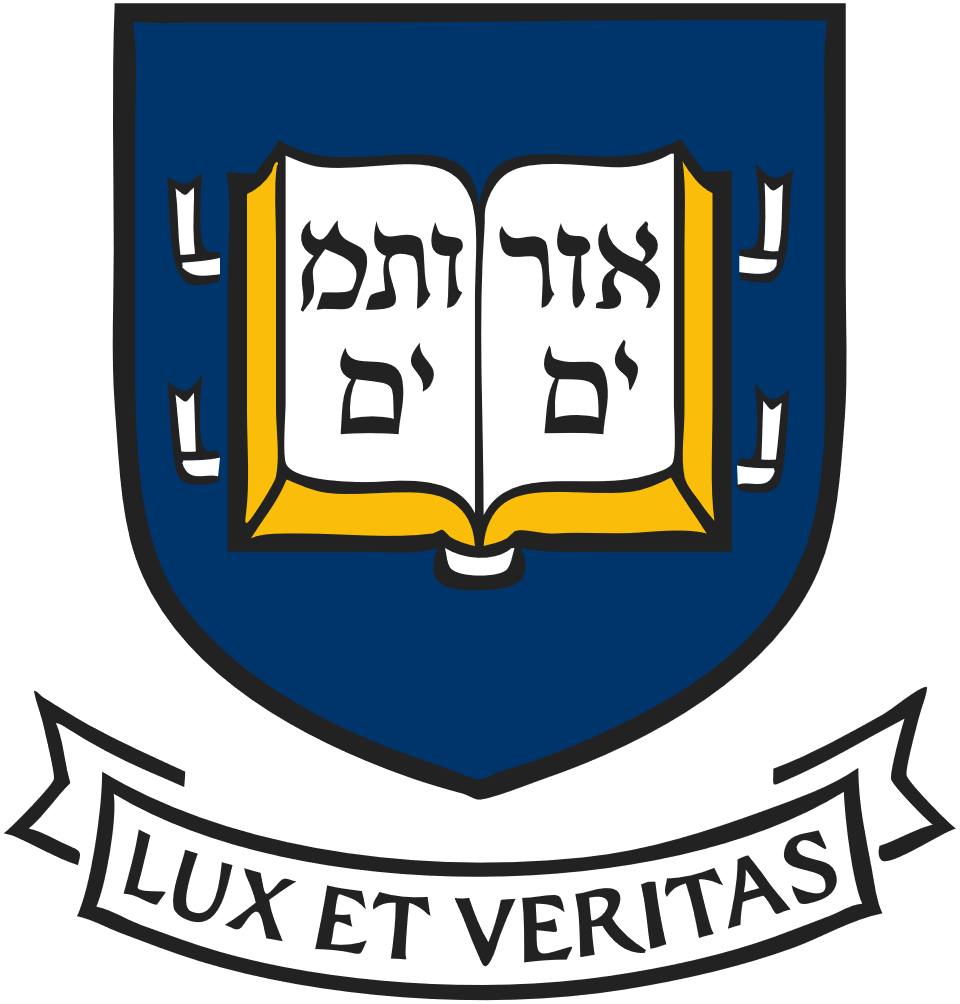}\hspace{0.6em}
			\includegraphics[height=1.0cm]{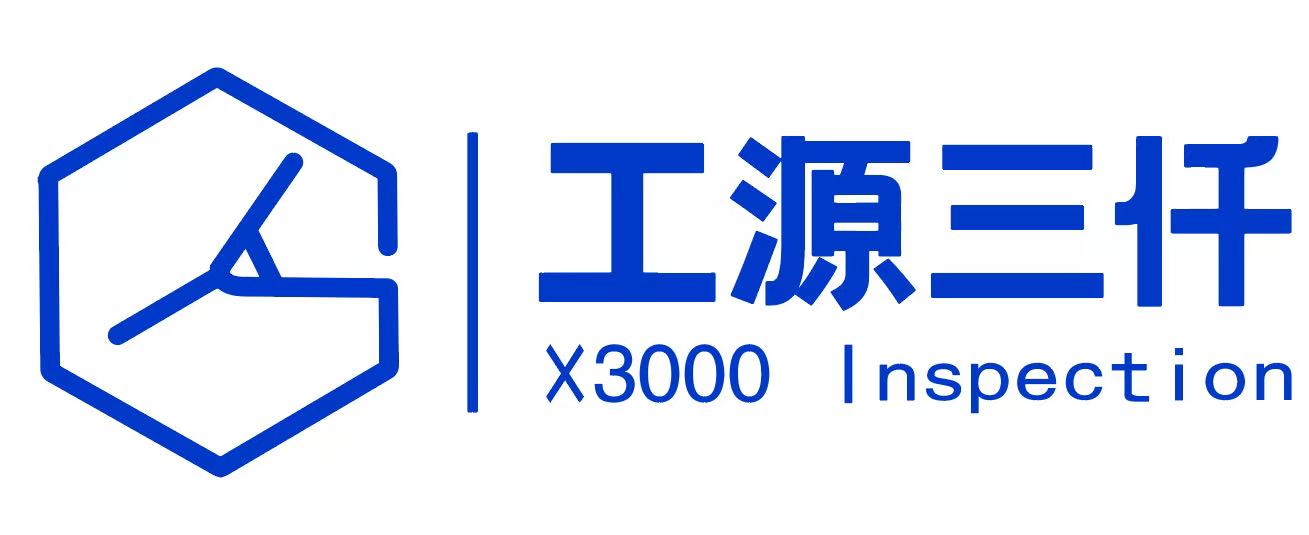}\hspace{0.1em}% 学校2
		\end{minipage}
		\\[0.4em]
		\textcolor{academicblue}{\rule{\textwidth}{0.6pt}}%
	\end{minipage}%
}

\settitlepreskip{-10ex}
\settitlelogogap{3em}

\showlogointitle
\title{Hard, Yet Reducible: Controlled Forward Transfer for Synthetic Degradation Curation}
\shorttitle{CuRB}

\makeatletter

\newcommand{\manualtitlecontent}{%
	\begin{center}
	{\MA@author@font
		Chunming He$^{1,2}$,
		Kailai Zhou$^{3}$,
		Jiaming Zuo$^{1,2}$,
		Hanqi Liu$^{3,2}$,
		Fengyang Xiao$^{1}$,
			\par\vspace{0.4em}
		Youwei Pang$^{2}$,
		Xiaofeng Liu$^{4}$,
		Weisi Lin$^{3}$,
		Xiaoqi Zhao$^{3}$
		\par}
	
	\vspace{\MA@author@affiliation@skip}
	
	{\MA@affiliation@font
		$^{1}$Tsinghua University, China\par
		$^{2}$X3000 Inspection Co., Ltd\par
		$^{3}$AI4X Team, Nanyang Technological University, Singapore\par
		$^{4}$Yale University, USA\par
	}
	
	\vspace{\MA@affiliation@email@skip}
	
\end{center}

	\vspace{\MA@author@abstract@preskip}
	{\centering\textcolor{abstractborder}{\rule{0.3\textwidth}{0.4pt}}\par}
	\vspace{\MA@author@abstract@postskip}

	\noindent
Selecting synthetic degradations for dense prediction requires an estimate of their training utility, the generalization gain they bring under a finite training budget.
Clean and degraded twins share content and labels, suggesting a score based on how much short training reduces the excess error caused by degradation.
However, this gap can also shrink when clean performance deteriorates.
Measuring the improvement on degraded images alone avoids that confound, but it still credits progress that the same amount of clean training would have produced.
We propose the \textbf{controlled Reducible Degradation Gap} (cRDG) for regions defined by degradation type and severity.
From a common checkpoint, cRDG runs two budget-matched probes that differ only in one augmentation slot, which holds either a synthetic degradation or a clean augmentation.
The score is the gain on held-out degraded images relative to the clean-control probe.
Clean harm is a separate feasibility constraint.
cRDG reveals a correctable severity band in which training on the degradation yields high controlled gain under the available budget, and the band moves with the predictor, the starting checkpoint, and the training budget.
\textbf{Curation of Reducible Bands} (\method) uses cRDG to select synthetic data without changing the predictor.
On semantic segmentation and salient object detection, \method{} improves representative predictors under matched synthetic-data budgets and training schedules, extends to existing data-generation pipelines, and preserves clean performance.
Code and supporting materials will be publicly released.
}

\renewcommand{\@abstractboxcontent}{\manualtitlecontent}

\makeatother
\begin{document}
\maketitle

\section{Introduction}
\label{sec:intro}

\looseness=-1
{Generators can now produce unlimited variants of labelled data.
	The binding question is no longer how to synthesize but what to spend a finite training budget on~\citep{he2023strategic,he2026refining}.
	Under a fixed schedule, every degraded sample occupies a slot that could hold a clean augmentation instead.
	We take this alternative as the reference, and the value of a degradation is the extra generalization from degradation.}

\looseness=-1
{Data selection often conflates \emph{visual quality}, \emph{instantaneous difficulty}, and \emph{training utility}.
	The first two describe the current sample or the model's response to it, whereas training utility is the gain obtained by actually training on that sample.
	Utility is therefore the quantity relevant to selection, and it depends on the model state, the optimizer, and the training budget.
	We study this quantity for synthetic degradation in dense prediction, where degraded variants of clean labelled images keep scene content, geometry, and labels fixed~\citep{rombach2022high,zhang2023adding,rajagopalan2025gendeg}.}

Higher visual fidelity does not guarantee higher downstream utility, and recent studies attribute this mismatch to limited support and diversity in generated data~\citep{adamkiewicz2026pretty,zhang2026makes}.
Within a plausible candidate pool, realism acts as a support constraint rather than a ranking criterion.
Difficulty is also a poor indicator of utility, because a strong degradation can remove the evidence needed to recover the mask, leaving samples that are hard yet carry little learnable signal under the available budget.
Selecting by loss or uncertainty~\citep{mindermann2022prioritized,askari2025improving} can then waste the budget or cause negative transfer.

\begin{figure}[t!]
	\centering
	\includegraphics[width=\linewidth]{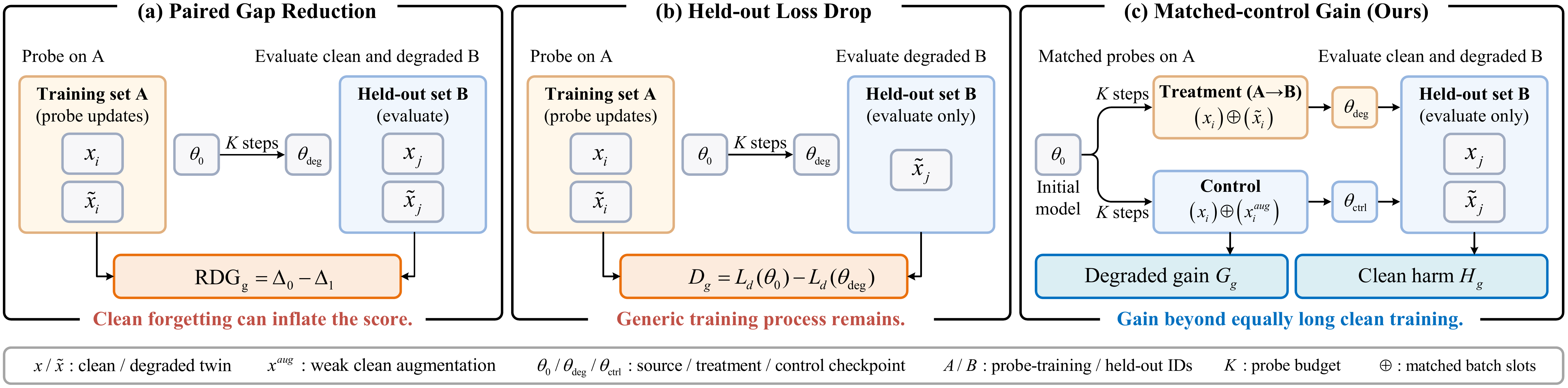}
	\caption{\textbf{Why Matched Clean Control Matters.}
		(a)~Paired gap reduction can be inflated by clean forgetting.
		(b)~Held-out degraded loss drop avoids this inflation but still credits generic training progress.
		(c)~Our matched-control formulation underlying \crdg{} measures degraded gain $G_g$ beyond equally long clean training, while checking clean harm $H_g$ separately.}
	\label{fig:concept}
	\vspace{-3mm}
\end{figure}

We define the utility of a degradation as its held-out gain beyond equally long clean training under the same probe budget, subject to a clean-retention constraint.
Forward transfer here means the gain on held-out source scenes that carry the same degradation after a short probe, and transfer to real adverse domains is evaluated separately (\Cref{sec:exp_score}).
Curation uses only source images and their synthetic twins, never target-domain observations.
We group candidates into \emph{regions} $g$ defined by degradation type and severity bin, so the policy chooses kinds of degradation rather than individual scenes.
Full retraining for every region would be costly, so short probes estimate held-out gains per region.
Synthetic degradation also offers a natural control, since each valid degraded image has a clean twin with the same scene and label.
Existing valuation scores, such as influence and gradient alignment~\citep{pruthi2020estimating,zhu2024generative}, holdout-loss prioritization~\citep{mindermann2022prioritized}, early-training statistics~\citep{paul2021deep,toneva2019forgetting}, and learnability-guided generation~\citep{chan2026learnability,askari2025improving}, do not combine this twin structure with a clean-control stream and held-out transfer at the region level.

This pairing suggests comparing degraded and clean loss for the same content.
Let $L_d(\theta_t)$ and $L_c(\theta_t)$ be the mean losses on the degraded twins of a region and on their clean counterparts at checkpoint $\theta_t$.
The gap $\Delta_t=L_d(\theta_t)-L_c(\theta_t)$ is the excess error caused by the degradation.
One could score a region by how much a short probe shrinks this gap:
\begin{align}
\Delta_0-\Delta_1 &= \underbrace{\left[L_d(\theta_0)-L_d(\theta_1)\right]}_{\text{degraded loss falls}}-\underbrace{\left[L_c(\theta_0)-L_c(\theta_1)\right]}_{\text{clean loss falls}}.
\label{eq:confound}
\end{align}
The second bracket exposes the problem.
A probe that forgets clean scenes also shrinks the gap, so clean forgetting inflates the score (\Cref{prop:confound}).
Dropping the clean term and scoring only the fall in degraded loss removes this inflation.
It still credits progress that equally long clean training would have produced anyway.
The quantity we need is the gain \emph{beyond} that clean-control trajectory (\Cref{fig:concept}).

We estimate this gain with matched treatment and clean-control probes.
Both start from the same checkpoint and differ only in whether the designated augmentation slot contains a degradation or a clean augmentation.
They are evaluated on held-out twins, and the two halves are swapped to obtain the \textbf{controlled Reducible Degradation Gap} (\crdg).
Scores stay signed so negative transfer remains visible, and a separate clean-harm constraint excludes regions that damage clean performance.

\crdg{} shows that utility is not monotonic in severity.
Each degradation type has a \emph{correctable band} in which training on the degradation yields high controlled gain under the budget, and the severity centroid of this band shifts with probe budget $K$, starting checkpoint $\theta_0$, and predictor setting, and differs across degradation types.
Per-region retraining at three final-training budgets shows the same shift in real-target gains, and proportionally scaled probes track it (\Cref{sec:exp_band}).
A fixed middle-severity rule (Fixed-Mid) serves as the sanity check for this claim throughout the experiments.

Around \crdg{} we build \textbf{Curation of Reducible Bands} (\method), a curation framework that adds degradation-aware selection to existing dense predictors without changing their architectures.
Scores are recomputed for the operating checkpoint and budget.
The band only summarizes the profile.
Selection ranks feasible positive-score regions directly and fills a fixed synthetic budget in rank order, with validity gating before ranking and $k$-center sampling for the last partial region.

Our contributions are as follows.
\begin{itemize}[leftmargin=*,itemsep=0pt,topsep=-\parskip,partopsep=0pt,parsep=0pt]
	
	\item We formulate \emph{budget-conditioned correctability}.
	Training utility need not track severity, realism, or difficulty, and the useful severity region migrates with predictor, checkpoint, and probe budget.
	\item We propose \crdg, a held-out, budget-matched estimate of the gain from training on a degradation region relative to an equally long clean-control probe, with clean harm handled separately.
	\item We introduce \method{}, a plug-and-play framework that adds degradation-aware curation to existing dense predictors without changing their architectures.
	\item Under a matched protocol, \method{} outperforms region-level adaptations of existing selection methods.
	Across SemSeg and SOD, its gains extend to the evaluated data-generation settings while preserving clean performance.
\end{itemize}

\section{Related Work}
\label{sec:related}

\textbf{Synthetic Degradation and Degradation-Aware Training.}
Diffusion-based methods synthesize degradation patterns and weather conditions~\citep{rajagopalan2025gendeg,nguyen2026cyclone}.
Restoration-driven methods~\citep{rass2026} inject enhancement knowledge into downstream segmentation.
Synthesis frameworks such as DGInStyle~\citep{jia2023dginstyle} and CA-LoRA~\citep{park2026calora} generate labelled training images for segmentation.
These approaches decide \emph{how} training data are generated or used.
We ask \emph{which} synthetic degradations deserve a finite training budget.
Higher fidelity alone does not guarantee training value.
\citet{adamkiewicz2026pretty} link the mismatch to limited coverage and diversity, and \citet{zhang2026makes} show that scene composition and instance fidelity matter for segmentation.
We instead hold content and labels fixed and vary only degradation type and severity, which allows a controlled estimate of degradation-specific utility.
\method{} is a curation layer between an existing generator and an existing predictor and changes neither.
We evaluate it with ISSA~\citep{li2023issa}, procedural weather corruption~\citep{hendrycks2019benchmarking}, and Gen4Seg~\citep{yin2026gen4seg} under their original settings.
UDA methods~\citep{hoyer2022daformer,tranheden2021dacs,sakaridis2019darkzurich} instead use unlabelled target images during adaptation.
Our setting uses only source labels and synthetic twins during curation and training.

\textbf{Data Valuation and Learnability-Guided Selection.}
Influence- and gradient-based methods~\citep{pruthi2020estimating,zhu2024generative} estimate how much a sample contributes to learning.
RHO-Loss~\citep{mindermann2022prioritized} prioritizes examples by their effect on holdout loss.
Early-training statistics such as EL2N, GraNd, and forgetting events~\citep{paul2021deep,toneva2019forgetting} find informative examples from short trajectories.
Learnability-guided generation~\citep{chan2026learnability,askari2025improving} steers synthesis with learner feedback.
Our setting differs in that utility is estimated at the region level using content-matched twins, a clean-control probe, and held-out transfer under a fixed budget (\Cref{sec:crdg}).
Diffusion Curriculum~\citep{discl2025} is closest in spirit and adapts the guidance level of generated data using validation feedback.
It does not compare against an equally long clean-control probe, which is the counterfactual our score is built on.

\section{Controlled Reducibility for Degradation Curation}
\label{sec:method}

\begin{figure}[t!]
	\centering
	\includegraphics[width=\linewidth]{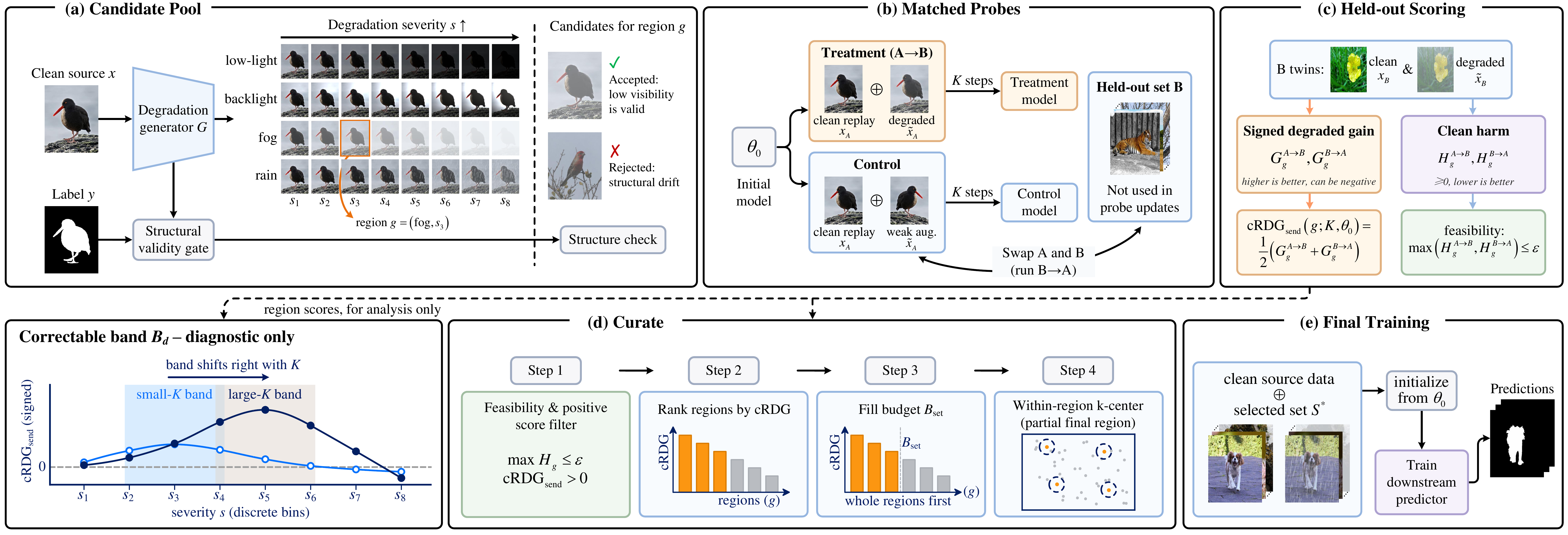}
	\caption{\textbf{Overview of \method{}.}
		(a)~Degradation-severity regions are generated and validity-filtered.
		(b)~Two matched probes start from $\theta_0$ and differ only in the augmentation slot, where $\oplus$ denotes the union of the clean-replay stream and the slot stream.
		(c)~Held-out evaluation yields degraded gain $G_g$ and clean harm $H_g$. Averaging $G_g$ over the two split directions gives \crdg{}. $H_g$ determines feasibility, while the correctable band is diagnostic only.
		(d)~Feasible positive-score regions are ranked and fill the budget $B_{\mathrm{sel}}$ in rank order, with $k$-center sampling for the last partial region.
		(e)~Final training restarts from $\theta_0$ on clean data, with $\mathcal{S}^\star$ in the synthetic slots under the fixed schedule.
	}
	\label{fig:overview}
	\vspace{-3mm}
\end{figure}

\method{} selects synthetic degradations for a given predictor and training budget without changing the predictor.
\Cref{fig:overview} shows the pipeline.
Matched probes estimate a region-level utility, \crdg{}, and selection ranks regions by \crdg{} under a clean-harm constraint and a fixed synthetic-data budget.

\subsection{Candidate Regions and Validity}
\label{sec:setup}

We are given a labelled clean source set $\mathcal{D}_s=\{(x_i,y_i)\}_{i=1}^{N}$.
A generator $\mathcal{G}$ produces degraded twins $\tilde{x}_{i,d,s}=\mathcal{G}(x_i;d,s,z)$ with degradation type $d$, severity level $s$, and stochastic detail $z$.
We use four appearance degradations that preserve object geometry after validity filtering.
Low light and backlight cover illumination changes, and fog and rain cover weather effects.
Each family is divided into eight ordered severity levels, giving $32$ regions.
This discretization resolves severity-dependent effects while keeping the probe set small.
A \emph{degradation region} $g=(d,s)$ contains candidates from one family and one severity bin, where ``region'' denotes a group in the candidate pool rather than a spatial image region.
\method{} ranks regions, not individual scenes.
The region definition is shared across architectures.
Its estimated utility is recomputed for each model and budget.

\textbf{Validity gate.}
\label{sec:gate}
Severe degradation does not by itself invalidate the mask.
We therefore separate label validity from reducibility.
A binary gate rejects candidates only when generation changes scene structure.
It checks frozen DINO feature correspondence near ground-truth boundaries and passes a uniformly darker image with intact geometry.
Details and the gate audit are in \Cref{app:gate}.

\textbf{The twin gap.}
Write $L_d^{S}(\theta;g)$ for the mean loss on the degraded twins of an index set $S$ and $L_c^{S}(\theta)$ for the mean loss on their clean images.
At the source checkpoint $\theta_0$, the gap $\Delta_0(g)=L_d^{S}(\theta_0;g)-L_c^{S}(\theta_0)$ isolates the error caused by the degradation, since twins share scene, geometry, and label.
It measures what is not yet mastered, rather than what can be mastered within a given budget.

\subsection{Matched-Control Probes}
\label{sec:confound}

Consider a short probe $\theta_0\!\to\!\theta_1$ on region $g$, scored by the shrinkage $\Delta_0-\Delta_1$.
By \Cref{eq:confound} this equals the drop in degraded loss minus the drop in clean loss, so it rewards clean forgetting:
\begingroup
\setlength{\topsep}{0pt}
\setlength{\partopsep}{0pt}
\begin{proposition}[Gap reduction can reward clean forgetting]
	\label{prop:confound}
	$\Delta_0-\Delta_1>0$ can hold even when the degraded loss worsens, $L_d(\theta_1)>L_d(\theta_0)$, whenever $L_c(\theta_1)-L_c(\theta_0)>L_d(\theta_1)-L_d(\theta_0)>0$.
\end{proposition}
\endgroup
Scoring the degraded-loss drop alone removes this inflation but still credits progress that equally long clean training would produce.
We therefore compare each degradation probe with an equally long clean-control probe.
We report the uncontrolled score as the baseline \rdg{}, defined together with \crdg{} in \Cref{sec:crdg}.

To measure transfer rather than fit to the probe samples, we split source image IDs once into halves $A$ and $B$, train on one half, and evaluate on the other (cross-fitting).
The held-out half is excluded from the current probe update, although it may have contributed to training $\theta_0$.
The split is reused for every region.
Let $U(\theta,\mathcal{S};K)$ denote $K$ optimizer steps from $\theta$ on a data stream $\mathcal{S}$, and let $\oplus$ denote the union of the clean-replay stream and the augmentation-slot stream in each probe batch.
From the common start $\theta_0$ we run two probes on half $A$,
\begin{equation}
\theta^{\mathrm{deg}}_{A,g}=U\!\big(\theta_0,\;A_{\mathrm{clean}}\oplus A_{\mathrm{deg},g};\,K\big),
\qquad
\theta^{\mathrm{ctrl}}_{A}=U\!\big(\theta_0,\;A_{\mathrm{clean}}\oplus A_{\mathrm{aug}};\,K\big).
\label{eq:branches}
\end{equation}
The treatment probe fills the slot with twins from region $g$.
The clean-control probe fills the same slot with a weak photometric augmentation of the same base view (\Cref{app:probe}).
Both probes share source IDs, steps, batch composition, clean-replay ratio, optimizer, schedule, data order, and initialization.
The control does not depend on $g$, so it is run once per direction and shared across all regions.
If no valid twin exists after $n_z$ regeneration draws, the control view is kept in the treatment slot, so the match is never broken.
Held-out twins that fail the gate are dropped from both evaluations.

\textbf{Operating rule for $K$.}
We set $K$ to $1/80$ of the final training schedule and scale it with the schedule, so the probe-to-training budget ratio stays fixed across operating points.

\subsection{Matched-Control Reducibility}
\label{sec:crdg}

Both probes are evaluated on the held-out half $B$.
We define the \emph{degraded gain} and the \emph{clean harm} as
\begin{equation}
G^{A\to B}_g=L_d^{B}\!\big(\theta^{\mathrm{ctrl}}_{A};g\big)-L_d^{B}\!\big(\theta^{\mathrm{deg}}_{A,g};g\big),
\qquad
H^{A\to B}_g=\Big[L_c^{B}\!\big(\theta^{\mathrm{deg}}_{A,g}\big)-L_c^{B}\!\big(\theta^{\mathrm{ctrl}}_{A}\big)\Big]_+ .
\label{eq:harm}
\end{equation}
$G$ is positive when degradation training outperforms the matched clean control on held-out degraded images.
$H$ is positive when the treatment increases the held-out clean loss.
We then swap the two halves and average the two held-out gains,
\begin{equation}
\crdgs(g;K,\theta_0)=\tfrac12\big(G^{A\to B}_g+G^{B\to A}_g\big).
\label{eq:crdg}
\end{equation}
A region is \emph{feasible} iff $\max\big(H^{A\to B}_g,H^{B\to A}_g\big)\le\epsilon$.
Infeasible regions are excluded from selection, and signed scores remain available for analysis.
We use a hard constraint rather than a penalty $G_g-\lambda H_g$ because it enforces clean retention without another trade-off parameter.
The constraint bounds probe-time harm and does not guarantee final clean accuracy, which \Cref{sec:exp_main} checks empirically.

\textbf{Baselines from the same probes.}
The same held-out losses define two baselines.
The paired gap reduction is $\rdg^{A\to B}_g=[L_d^{B}(\theta_0;g)-L_d^{B}(\theta^{\mathrm{deg}}_{A,g};g)]-[L_c^{B}(\theta_0)-L_c^{B}(\theta^{\mathrm{deg}}_{A,g})]$, and the held-out loss drop is $D^{A\to B}_g=L_d^{B}(\theta_0;g)-L_d^{B}(\theta^{\mathrm{deg}}_{A,g};g)$.
Like $G_g$, both are averaged over the directions $A\to B$ and $B\to A$.
$D_g$ differs from \crdg{} only in what is subtracted, the initial loss instead of the clean-control loss, so it is the direct ablation of the control branch, whereas \rdg{} additionally credits clean forgetting.
\crdg{} is a short-horizon, source-side estimate of controlled degradation utility.
Its relation to full-training gains on real target domains is evaluated empirically in \Cref{sec:exp_score}.

\subsection{Curation and Final Training}
\label{sec:train}

\textbf{Selection.}
\method{} first removes infeasible and non-positive regions.
The remaining regions are ranked by \crdg{} and filled in descending order until $B_{\mathrm{sel}}$ is reached.
Each region is exhausted before moving to the next.
If only part of the final region is needed, we use feature-space $k$-center sampling.
Any unfilled slots revert to clean data, so the total number of training samples is unchanged.

\textbf{Final objective.}
The curation procedure can be used with different final-training objectives, and all matched-curation methods share the one below.
Starting again from $\theta_0$, we train on clean data plus $\mathcal{S}^\star$ with
\begin{equation}
\mathcal{L}=\mathcal{L}_{\mathrm{seg}}(f(x_i),y_i)
+\lambda_s\,\mathcal{L}_{\mathrm{seg}}(f(\tilde{x}_i),y_i)
+\lambda_c\!\sum_p w_i(p)\,\mathrm{KL}\!\big(f(x_i)_p\,\|\,f(\tilde{x}_i)_p\big).
\label{eq:trainloss}
\end{equation}
Here $w_i(p)=M_i(p)\,(1+\lambda_b B_{y_i}(p))$, $B_{y_i}(p)$ marks a ground-truth boundary band, and $M_i(p)$ is the cosine similarity between the clean DINOv2 feature at $p$ and the degraded feature at its matched location $q_p$, mapped linearly to $[0,1]$.
The KL term compares class-probability maps.
Matched-curation experiments use \Cref{eq:trainloss} unchanged, and plug-in experiments keep each host's own objective.

\textbf{Cost.}
With $M$ regions and $R$ split directions, curation costs $MR$ treatment probes plus $R$ shared controls.
Changing the predictor or the schedule requires recomputing \crdg{} but never changes the architecture.
\Cref{sec:exp_cost} reports compute matched between \method{} and the baselines.

\subsection{The Diagnostic Severity Band}
\label{sec:band}

\crdg{} depends on the probe budget and the checkpoint, not on severity alone.
For a type $d$ with bins $s_1<\dots<s_{N_s}$, let $\mathcal{F}_d$ contain the feasible bins with positive \crdg{}.
The \emph{correctable band} is the set of bins whose score is at least half of the family maximum,
\begin{equation}
\mathcal{B}_d(K,\theta_0)=\big\{s\in\mathcal{F}_d:\crdgs\big((d,s);K,\theta_0\big)\ge\kappa\max\nolimits_{s'\in\mathcal{F}_d}\crdgs\big((d,s');K,\theta_0\big)\big\},
\label{eq:band}
\end{equation}
with $\kappa=1/2$.
For smoother comparison across operating conditions, we also report the \crdg-weighted severity centroid $\bar{s}_d(K,\theta_0)=\sum_{s\in\mathcal{F}_d}s\,\crdgs((d,s);K,\theta_0)\big/\sum_{s\in\mathcal{F}_d}\crdgs((d,s);K,\theta_0)$, which is undefined when $\mathcal{F}_d$ is empty.
Severity is represented by normalized bin centers $s_k=(2k-1)/16$, $k=1,\ldots,8$.
Both are diagnostics, and selection ranks regions directly by \crdg{}.
The \textbf{band-migration} hypothesis is that $\bar{s}_d$ shifts with the probe budget $K$, the checkpoint, and the predictor, and differs across types.
\Cref{sec:exp_band} tests whether the real band shifts with the training budget and whether scaled-$K$ \crdg{} tracks it.

\section{Experiments}
\label{sec:exp}

\begin{table}[t]
	\setlength{\abovecaptionskip}{1mm}
	\centering
	\begin{subtable}[t]{0.49\linewidth}
		\centering
		\resizebox{\linewidth}{!}{
			\setlength{\tabcolsep}{1.7mm}
			\begin{tabular}{lcccc}
				\toprule
				Method & $M\downarrow$ & $F_\beta^{\max}\uparrow$ & $E_\phi\uparrow$ & $S_\alpha\uparrow$ \\
				\midrule
				WFANet~\citep{wxsod2025}             & .016 & .908 & .953 & .925 \\
				\quad + Uncurated twins ($B_{\mathrm{sel}}$)    & .015 & .912 & .956 & .926 \\
				\rowcolor{rowhl}
				\quad + \method{} ($B_{\mathrm{sel}}$)          & \textbf{.013} & \textbf{.923} & \textbf{.963} & \textbf{.933} \\
				CSSAM~\citep{cong2025breaking}           & .011 & .937 & .972 & .940 \\
				\quad + Uncurated twins ($B_{\mathrm{sel}}$)    & .011 & .939 & .975 & .942 \\
				\rowcolor{rowhl}
				\quad + \method{} ($B_{\mathrm{sel}}$)          & \textbf{.009} & \textbf{.952} & \textbf{.987} & \textbf{.947} \\
				NUN~\citep{he2026nested}             & .014 & .920 & .965 & .933 \\
				\quad + Uncurated twins ($B_{\mathrm{sel}}$)    & .013 & .924 & .968 & .935 \\
				\rowcolor{rowhl}
				\quad + \method{} ($B_{\mathrm{sel}}$)          & \textbf{.011} & \textbf{.934} & \textbf{.978} & \textbf{.938} \\
				\bottomrule
		\end{tabular}}
		\caption{WXSOD-real~\citep{wxsod2025}.}
		\label{tab:bench_wxsod}
	\end{subtable}
	\hfill
	\begin{subtable}[t]{0.49\linewidth}
		\centering
		\resizebox{\linewidth}{!}{
			\setlength{\tabcolsep}{1.7mm}
			\begin{tabular}{lcccc}
				\toprule
				Method & $M\downarrow$ & $F_\beta^{\max}\uparrow$ & $E_\phi\uparrow$ & $S_\alpha\uparrow$ \\
				\midrule
				SI-EDN~\citep{li2024size}          & .062 & .817 & .841 & .810 \\
				\quad + Uncurated twins ($B_{\mathrm{sel}}$)    & .058 & .835 & .858 & .817 \\
				\rowcolor{rowhl}
				\quad + \method{} ($B_{\mathrm{sel}}$)          & \textbf{.052} & \textbf{.853} & \textbf{.873} & \textbf{.831} \\
				CSSAM~\citep{cong2025breaking}           & .035 & .887 & .916 & .886 \\
				\quad + Uncurated twins ($B_{\mathrm{sel}}$)    & .034 & .891 & .919 & .888 \\
				\rowcolor{rowhl}
				\quad + \method{} ($B_{\mathrm{sel}}$)          & \textbf{.030} & \textbf{.912} & \textbf{.934} & \textbf{.893} \\
				NUN~\citep{he2026nested}             & .038 & .857 & .896 & .875 \\
				\quad + Uncurated twins ($B_{\mathrm{sel}}$)    & .036 & .867 & .902 & .877 \\
				\rowcolor{rowhl}
				\quad + \method{} ($B_{\mathrm{sel}}$)          & \textbf{.032} & \textbf{.880} & \textbf{.917} & \textbf{.886} \\
				\bottomrule
		\end{tabular}}
		\caption{CSOD10K~\citep{cong2025breaking}.}
		\label{tab:bench_csod}
	\end{subtable}
	\caption{\textbf{Evaluation on SOD Benchmarks} following the official protocol (\Cref{sec:exp_setup}).
		At the same augmentation budget, comparing uncurated sampling with \method{} isolates the gain from curation.
	} 
	\vspace{-2mm}
	\label{tab:bench}
	% \vspace{-2mm}
\end{table}

\subsection{Setup}
\label{sec:exp_setup}

\textbf{Data and protocols.}
We use two core evaluation protocols.
For SOD, WXSOD~\citep{wxsod2025} and CSOD10K~\citep{cong2025breaking} follow their official splits and metrics, with each base model using its published training recipe.
Augmented variants use the same budget of degraded twins from clean DUTS-TR~\citep{wang2017duts} and the same final-training schedule (\Cref{app:sodprotocol}).
For SemSeg, we follow the Cityscapes${\to}$ACDC / Dark Zurich protocol of ISSA~\citep{li2023issa}, also used by Gen4Seg~\citep{yin2026gen4seg}.
Cityscapes~\citep{cordts2016cityscapes} is the only source, and SegFormer follows the released ISSA configuration (\Cref{app:impl}).
The matched synthetic pool contains low light, backlight, fog, and rain at eight severity levels per family.
Low light is aligned with the night condition, while fog and rain have direct counterparts in ACDC~\citep{sakaridis2021acdc}.
Backlight is an additional source-side illumination family, and snow is held out entirely for unseen-family evaluation.
We report the in-family mean over fog, night, and rain, and evaluation also includes Dark Zurich~\citep{sakaridis2019darkzurich} and Cityscapes val.
Plug-in experiments apply \method{} under each host's published setting and synthetic-data budget.
Operational curation uses one fixed probe seed and one fixed $A|B$ partition.
Two additional partitions are used only for stability analysis (\Cref{app:probe}).

\textbf{Comparison groups.}
\emph{Benchmark baselines}: WFANet, CSSAM, and NUN on WXSOD, and SI-EDN, CSSAM, and NUN on CSOD10K.
\emph{Data-generation hosts}: ISSA~\citep{li2023issa} with SegFormer and RobustNet~\citep{choi2021robustnet} / DeepLabV3+~\citep{chen2018deeplabv3plus}, Hendrycks-Weather~\citep{hendrycks2019benchmarking}, and Gen4Seg~\citep{yin2026gen4seg}.
\emph{Selection baselines}: realism (NR-IQA / CLIPScore), highest loss / uncertainty, RHO-Loss-style selection~\citep{mindermann2022prioritized}, gradient contribution~\citep{zhu2024generative}, and a DisCL-style severity preference~\citep{discl2025} (\Cref{app:full}).
\emph{Twin-based scores}: $\Delta_0$, uncontrolled $\rdg$, held-out loss drop $D_g$, and $\crdgs$.
\textbf{Fixed-Mid}, the middle severity bins of every family, is a sanity control.
Source-only and All-gen.\ are no-ranking anchors.
Full-pool exposure appears only in the compute analysis.

\begin{table}[t]
	\setlength{\abovecaptionskip}{1mm}
	\centering
	\resizebox{\linewidth}{!}{
		\setlength{\tabcolsep}{1.3mm}
		\begin{tabular}{ll cc cc c}
			\toprule
			& & \multicolumn{2}{c}{ACDC mIoU $\uparrow$} & \multicolumn{2}{c}{Dark Zurich mIoU $\uparrow$} & Cityscapes \\
			\cmidrule(lr){3-4}\cmidrule(lr){5-6}
			\multicolumn{1}{l}{\multirow{-2}{*}{Data-generation host}} & \multicolumn{1}{l}{\multirow{-2}{*}{Published host setting}}  & Original & \cellcolor{rowhl} +\method{} & Original &\cellcolor{rowhl} +\method{} & val $\Delta$ $\uparrow$ \\
			\midrule
			\multicolumn{7}{@{}l}{\emph{single-generator hosts}}\\
			ISSA~\citep{li2023issa}              & SegFormer & 52.45 &\cellcolor{rowhl} \textbf{53.92} & 27.39 &\cellcolor{rowhl} \textbf{28.76} &\cellcolor{rowhl} $+0.24$ \\
			H-Weather~\citep{hendrycks2019benchmarking} & SegFormer & 49.21 &\cellcolor{rowhl} \textbf{50.63} & 23.44 &\cellcolor{rowhl} \textbf{24.65} &\cellcolor{rowhl} $+0.36$ \\
			Gen4Seg~\citep{yin2026gen4seg}        & SegFormer & 51.04 &\cellcolor{rowhl} \textbf{52.39} & 25.63 &\cellcolor{rowhl} \textbf{27.28} &\cellcolor{rowhl} $+0.45$ \\
			\midrule
			\multicolumn{7}{@{}l}{\emph{composite DG host}}\\
			ISSA~\citep{li2023issa} & RobustNet (DeepLabV3+) & 47.55 &\cellcolor{rowhl} \textbf{48.73} & 23.09 &\cellcolor{rowhl} \textbf{24.35} &\cellcolor{rowhl} $+0.18$ \\
			\bottomrule
	\end{tabular}}
	\caption{\textbf{\method{} as a plug-in to existing data-generation settings.}
		Published references for ISSA, H-Weather, and RobustNet+ISSA are taken from \citet{li2023issa}.
		Gen4Seg follows \citet{yin2026gen4seg} (its Dark Zurich value is obtained with the released pipeline).
		Our ``+\method{}'' runs apply \method{} under the corresponding host setting and synthetic-data budget.
		The composite host stacks RobustNet~\citep{choi2021robustnet} with ISSA on DeepLabV3+.
		Host-specific region construction: \Cref{app:plugin}.
	}
	\vspace{-2mm}
	\label{tab:plugin}
\end{table}

\textbf{Matched protocol.}
All selectors use the same gated pool, $B_{\mathrm{sel}}$, clean/degraded composition, initialization $\theta_0$, objective, and training schedule.
Sample-level baseline scores are averaged within each region before the common $k$-center step (\Cref{app:full}), so the comparison isolates region ranking.
Compute-matched baselines spend \method{}'s curation budget on extended final training (\Cref{sec:exp_cost}).

\begin{table}[t]
	\setlength{\abovecaptionskip}{1mm}
	\centering
	\resizebox{\linewidth}{!}{
		\setlength{\tabcolsep}{2.7mm}
		\begin{tabular}{l ccc c c c c}
			\toprule
			& \multicolumn{3}{c}{ACDC in-family} &in-fam. & Snow & Dark & Cityscapes \\
			\cmidrule(lr){2-4}
			\multicolumn{1}{l}{\multirow{-2}{*}{Method}} & Fog & Night & Rain & mean & (held out) &  Zurich & val \\
			\midrule
			Source-only                        & 60.61 & 28.42 & 50.35 & 46.46 & 48.79 & 24.08 & 67.84 \\
			\midrule
			\multicolumn{8}{@{}l}{\emph{matched curation protocol: same pool, gate, $B_{\mathrm{sel}}$, objective, and schedule}}\\
			All-gen.\ (uniform, $B_{\mathrm{sel}}$) & 63.82 & 31.51 & 52.78 & 49.37 & 50.15 & 26.55 & 68.21 \\
			Realism (NR-IQA / CLIPScore)       & 63.34 & 30.97 & 53.02 & 49.11 & 50.31 & 26.12 & 68.44 \\
			Highest loss / uncertainty         & 64.05 & 31.88 & 52.41 & 49.45 & 49.97 & 26.71 & 67.93 \\
			RHO-Loss style~\citep{mindermann2022prioritized} & 64.19 & 32.24 & 52.96 & 49.80 & 50.28 & 27.06 & 68.31 \\
			Grad.\ contrib.~\citep{zhu2024generative} & 64.73 & 31.85 & 53.42 & 50.00 & 50.55 & 26.84 & 68.58 \\
			DisCL-style selection~\citep{discl2025} & 64.48 & 32.36 & 53.27 & 50.04 & 50.79 & 27.02 & 68.40 \\
			$\rdg$ (uncontrolled) & 65.11 & 32.29 & 53.76 & 50.39 & 50.47 & 27.45 & 67.41 \\
			Fixed-Mid                          & 65.02 & 32.91 & 53.58 & 50.50 & 50.68 & 27.33 & 68.87 \\
			\rowcolor{rowhl}
			\method{} ($\crdgs$)               & \textbf{66.73} & \textbf{34.26} & \textbf{54.93} & \textbf{51.97} & \textbf{51.38} & \textbf{28.75} & \textbf{70.13} \\
			\bottomrule
	\end{tabular}}
	\caption{\textbf{Controlled curation comparison on Cityscapes${\to}$ACDC / Dark Zurich.} All methods share the candidate pool, validity gate, $B_{\mathrm{sel}}$, final-training objective, and schedule, differing in their sample-selection policies. Fog, night, and rain are treated as the in-family conditions. Snow is held out as an unseen family, with no snow samples entering the candidate pool, curation, or final training.}
	\vspace{-1mm}
	\label{tab:main}
\end{table}

\subsection{Main Results}
\label{sec:exp_main}

\textbf{Benchmark-level performance.}
\label{sec:exp_bench}
\Cref{tab:bench} attaches \method{} to three published base models on SOD benchmarks, including the SAM-based CSSAM, under corresponding protocols. The uncurated row adds the same budget of synthetic twins without selection and brings only small gains, with almost none on the strongest base model. Selecting the same budget with \crdg{} improves all four metrics on every base model and both benchmarks. The margin over uncurated twins remains positive throughout, including for CSSAM, where uncurated data provide little benefit.

\begin{figure}[t!]
\setlength{\abovecaptionskip}{1mm}
  \centering
  \includegraphics[width=\linewidth]{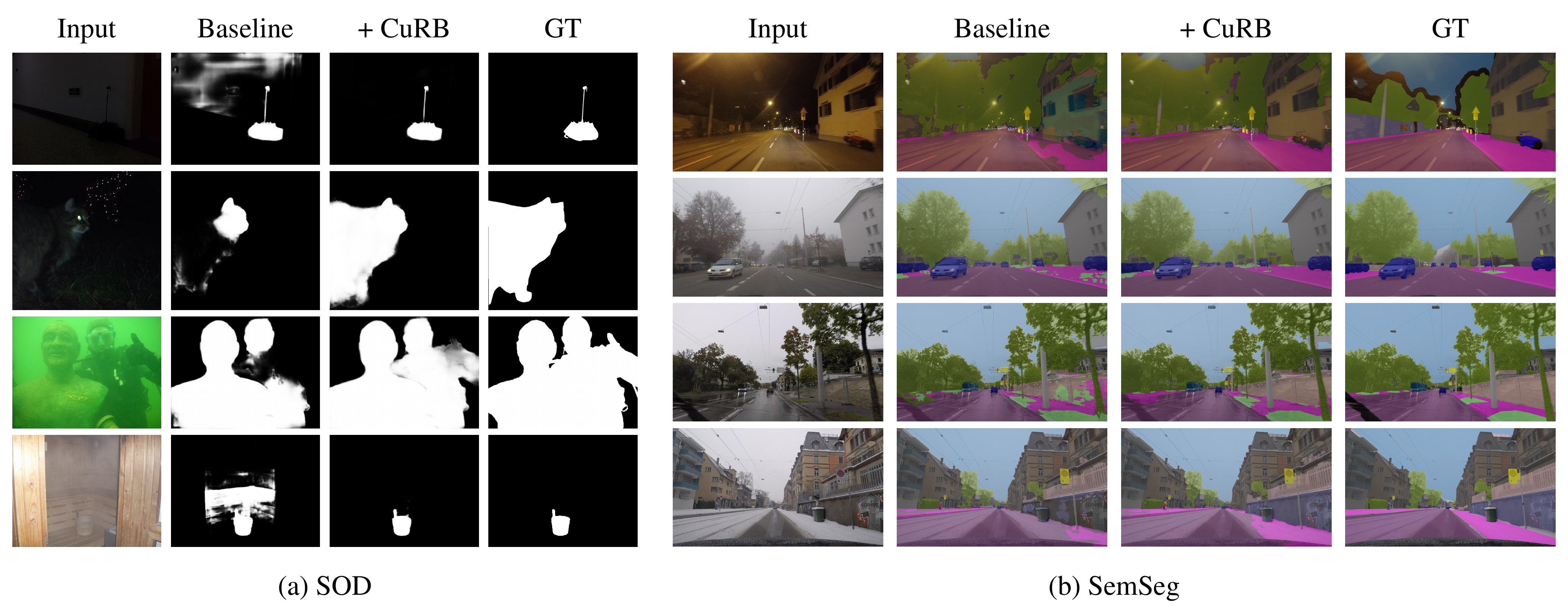}
\caption{\textbf{Qualitative comparison.}
(a)~Degraded SOD with NUN.
(b)~adverse-condition SemSeg with ISSA SegFormer.
Baseline and +\method{} predictions are shown for both tasks.}
\label{fig:qual}
  \vspace{-3mm}
\end{figure}

\textbf{Across data-generation frameworks.}
\label{sec:exp_plugin}
\Cref{tab:plugin} evaluates \method{} across several published data-generation settings at their original synthetic-data budgets. The ``+\method{}'' variants obtain $+1.2$ to $+1.5$ higher ACDC mIoU and $+1.2$ to $+1.7$ higher Dark Zurich mIoU than the corresponding originals, while maintaining or slightly improving Cityscapes performance. The gains span style synthesis (ISSA), procedural weather corruption (H-Weather), and diffusion-based editing (Gen4Seg), and persist for the composite RobustNet+ISSA host. The same insertion also improves SOD baselines in \Cref{tab:bench}. Qualitative examples for both tasks and hosts are shown in \cref{fig:qual}.

\begin{table}[t]
	\setlength{\abovecaptionskip}{1mm}
	\centering
	\begin{minipage}[t]{0.505\linewidth}
		\centering
		\resizebox{\linewidth}{!}{
			\setlength{\tabcolsep}{1.5mm}
			\renewcommand{\arraystretch}{1.03}
			\begin{tabular}{lccc}
				\toprule
				Variant & \makecell{in-fam.\\mIoU$\uparrow$} & \makecell{City\\val$\uparrow$} & \makecell{$\rho_{\mathrm{glob}}$ vs.\\$U_{\mathrm{real}}\uparrow$} \\
				\midrule
				\rowcolor{rowhl}
				\method{} (full)                      & \textbf{51.97} & \textbf{70.13} & \textbf{0.79} \\
				paired gap reduction ($\rdg$)         & 50.39 & 67.41 & 0.61 \\
				held-out loss drop $D_g$ & 51.02 & 68.72 & 0.70 \\
				in-sample (no cross-fitting)          & 49.21 & 69.58 & 0.44 \\
				scored on training twins (no held-out)  & 50.51 & 69.85 & 0.58 \\
				w/o clean-harm constraint ($H$ free)  & 50.68 & 67.18 & 0.79 \\
				\bottomrule
		\end{tabular}}
		\caption{\textbf{Core ablations} (Cityscapes${\to}$ACDC). \rdg{} and $D_g$ replace \crdg{} by the paired gap reduction and the held-out loss drop. Other rows remove cross-fitting, held-out transfer, and the clean-harm constraint. $\rho_{\mathrm{glob}}$ is over all $32$ regions.}
		\label{tab:ablation}
	\end{minipage}
	\hfill
	\begin{minipage}[t]{0.478\linewidth}
		\centering
		\resizebox{\linewidth}{!}{
			\setlength{\tabcolsep}{1.5mm}
			\renewcommand{\arraystretch}{1.03}
			\begin{tabular}{lcccc}
				\toprule
				Methods & $M\downarrow$ & $F_\beta^{max}\uparrow$ & $E_\phi\uparrow$ & $S_\alpha\uparrow$ \\
				\midrule
				Source-only                       & .021 & .892 & .960 & .920 \\
				All-gen.\ (uniform, $B_{\mathrm{sel}}$) & .022 & .889 & .958 & .918 \\
				Grad.\ contrib.~\citep{zhu2024generative} & .022 & .890 & .959 & .919 \\
				$\rdg$ (uncontr.) & .023 & .885 & .955 & .914 \\
				Fixed-Mid                         & .021 & .891 & .960 & .919 \\
				$\crdgs$, $H$ unconstrained       & .025 & .879 & .950 & .908 \\
				\rowcolor{rowhl}
				\method{}                         & .021 & .893 & .961 & .921 \\
				\bottomrule
		\end{tabular}}
		\caption{\textbf{Clean retention on SOD} (DUTS-TE, NUN/PVTv2-B2). Source-only trains on clean DUTS-TR. Augmented variants add the same synthetic budget selected by each method, so any drop is the clean cost of that selection.}
		\label{tab:sodclean}
	\end{minipage}
	\vspace{-3mm}
\end{table}

\textbf{Controlled data-selection comparison.}
\label{sec:exp_select}
\Cref{tab:main} compares selection criteria under one shared recipe, so the differences reflect what is selected. \method{} is best on every column. It reaches $51.97$ in-family mIoU versus $50.50$ for the strongest control and $50.04$ for the strongest literature baseline, $28.75$ on Dark Zurich, and $51.38$ on held-out snow, where improvement reflects cross-family transfer. Cityscapes performance is also preserved, reaching $70.13$ versus $67.84$ for source-only. The uncontrolled twin score is the only selector below the source-only model. \Cref{sec:exp_analysis} examines the ranking quality behind these differences. Published augmentation and adaptation results are not mixed into this table because their training exposure differs from our held-out-family protocol. Some use synthetic snow or target images. \Cref{tab:plugin} instead compares \method{} within each host's native setting.

\textbf{Matched compute.}
\label{sec:exp_cost}
With $K=1/80$ of the final schedule, the $66$ probes cost $0.59\times$ one final-training run, or $1.59\times$ with final training. Candidate generation, held-out scoring, and selection are excluded. At matched probe-plus-training compute, All-gen.\ and Fixed-Mid reach $49.86$ and $51.02$ in-family mIoU, versus $51.97$ for \method{}. Full-pool training at $3.1\times$ reaches only $50.02$ (\Cref{app:cost}).

\subsection{Ablation Study}
\label{sec:exp_ablation}

The ablations isolate the main components of \crdg{} (\Cref{tab:ablation}). Replacing \crdg{} with paired gap reduction lowers $\rho_{\mathrm{glob}}$ from $0.79$ to $0.61$, while removing only the clean-control subtraction ($D_g$) lowers it to $0.70$, showing the value of both the matched control and the controlled difference. In-sample estimation and training-twin scoring further weaken the ranking, supporting held-out cross-fitted evaluation. Removing the $H$ constraint leaves $\rho_{\mathrm{glob}}$ unchanged at $0.79$ but drops Cityscapes accuracy from $70.13$ to $67.18$, indicating that reducibility and clean feasibility capture different aspects of the selection problem. \Cref{tab:sodclean} shows the same effect on SOD: \method{} preserves clean DUTS-TE performance, whereas the $H$-free variant falls on every metric, including $S_\alpha$ ($.921\to.908$).

\subsection{Analysis: Training Utility and the Correctable Band}
\label{sec:exp_analysis}

\begin{figure}[t]
	\centering
	\setlength{\abovecaptionskip}{1mm}
	\includegraphics[width=\linewidth]{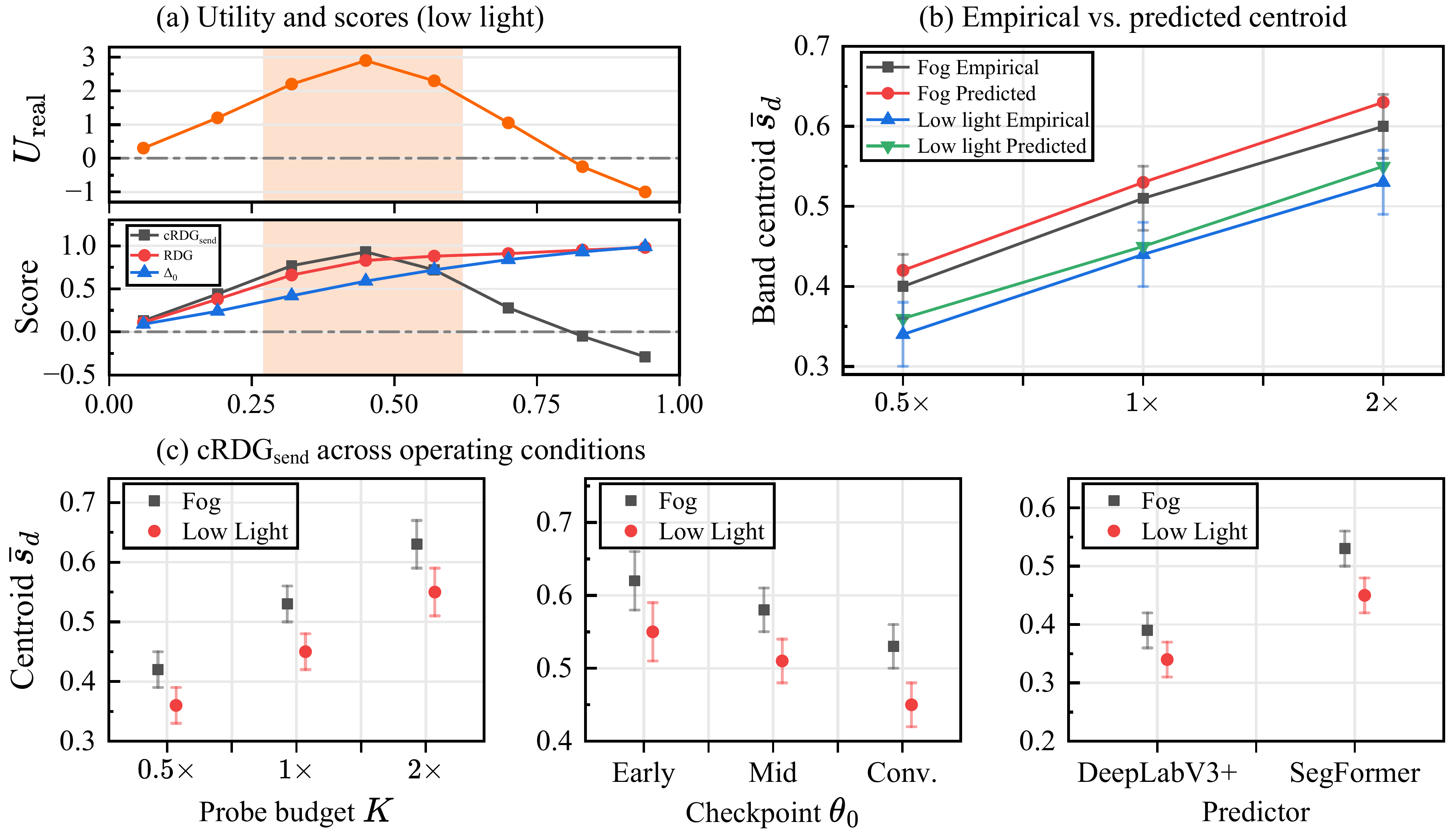}
	\caption{\textbf{Training utility and the correctable band.} (a) Per-bin retraining gains (top) and scores (bottom) for low light. Shading marks the predicted band. (b) Empirical severity centroids at three final budgets with scaled-$K$ $\crdgs$ predictions overlaid (paired-bootstrap $95\%$ CI). (c) $\crdgs$ centroids across probe budgets, checkpoints, and predictors. Error bars are s.d.\ over three partitions.}\vspace{-3mm}
	\label{fig:evidence}
\end{figure}

\textbf{Does \crdg{} predict real utility?}
\label{sec:exp_score}
We first obtain an empirical retraining utility. Holding everything but the region fixed, we retrain once per region $g$ and record the \emph{signed}, unnormalized gain on ACDC in-family, $U_{\mathrm{real}}(g)=\mathrm{mIoU}_{\mathrm{real}}(M_g)-\mathrm{mIoU}_{\mathrm{real}}(M_{\mathrm{clean}})$. \cref{fig:evidence}a overlays it with $\Delta_0$, \rdg, and $\crdgs$ for one family. 
We test whether \rdg{} remains high when real utility becomes negative and whether $\crdgs$ preserves the ordering of $U_{\mathrm{real}}$, including regions with negative utility.
The decision unit and the correlation unit are both the \emph{region} (per-sample real gain would need one retraining per sample). A global rank correlation can be inflated by mean differences \emph{between} families, so \Cref{tab:scores} reports $\rho_{\mathrm{glob}}$ over all $32$ regions and $\rho_{\mathrm{macro}}=\frac{1}{|\mathcal{D}|}\sum_d\rho_d$, where $\rho_d$ is the Spearman correlation over the $8$ regions of family $d$. Both use the operational ranking. 
Because \crdg{} is computed on source twins only, the table also correlates the same ranking with $U_{\mathrm{DZ}}$, the per-region retraining gain on Dark Zurich. Correlation with a second real target supports transfer beyond the benchmark used to define $U_{\mathrm{real}}$.
The held-out loss drop $D_g$ is the no-subtraction counterpart, and \crdg{} improves $\rho_{\mathrm{glob}}$ from $0.70$ to $0.79$ and in-family mIoU from $51.02$ to $51.97$ over it, and $\rho_{\mathrm{DZ}}$ from $0.62$ to $0.72$.

\begin{table}[t]
	\centering
	\resizebox{\linewidth}{!}{
		\setlength{\tabcolsep}{0.8mm}
		\renewcommand{\arraystretch}{1.05}
		\begin{tabular}{l ccc ccc >{\columncolor{rowhl}}c}
			\toprule
			& \multicolumn{3}{c}{Adapted valuation baselines}
			& \multicolumn{4}{c}{Twin-based scores} \\
			\cmidrule(lr){2-4}\cmidrule(lr){5-8}
			\multicolumn{1}{l}{\multirow{-2}{*}{Metric}}
			& \makecell{Highest loss /\\uncertainty}
			& \makecell{RHO-Loss style\\\citep{mindermann2022prioritized}}
			& \makecell{Grad. contrib.\\\citep{zhu2024generative}}
			& \makecell{$\Delta_0$\\(gap)}
			& \makecell{$\rdg$\\(uncontrolled)}
			& \makecell{$D_g$\\(held-out drop)}
			& \makecell{\method{}\\($\crdgs$)} \\
			\midrule
			$\rho_{\mathrm{glob}}$ with $U_{\mathrm{real}}$ $\uparrow$
			& 0.25 & 0.49 & 0.52 & 0.39 & 0.61 & 0.70 & \textbf{0.79} \\
			$\rho_{\mathrm{macro}}$ with $U_{\mathrm{real}}$ $\uparrow$
			& 0.17 & 0.35 & 0.37 & 0.27 & 0.43 & 0.56 & \textbf{0.74} \\
			$\rho_{\mathrm{DZ}}$ with $U_{\mathrm{DZ}}$ $\uparrow$
			& 0.21 & 0.43 & 0.45 & 0.33 & 0.53 & 0.62 & \textbf{0.72} \\
			in-fam.\ mIoU $\uparrow$
			& 49.45 & 49.80 & 50.00 & 49.28 & 50.39 & 51.02 & \textbf{51.97} \\
			\bottomrule
	\end{tabular}}
	\caption{\textbf{Region-level score comparison} under matched pool, gate, diversity, budget, objective, and schedule (Cityscapes${\to}$ACDC, ISSA SegFormer). $\rho_{\mathrm{glob}}$ is Spearman correlation with $U_{\mathrm{real}}$ over all $32$ regions, $\rho_{\mathrm{macro}}$ averages within-family correlations, and $\rho_{\mathrm{DZ}}$ evaluates the same ranking against per-region retraining gains on Dark Zurich. $\Delta_0$ is the unmastered gap and $D_g$ the held-out loss drop.}\vspace{-3mm}
	\label{tab:scores}
\end{table}

\textbf{How does the band move?}
\label{sec:exp_band}
\cref{fig:evidence}c examines variation with probe budget $K$, source checkpoint $\theta_0$, and predictor setting. Doubling $K$ shifts the centroid by about $0.10$, well beyond the $0.03$--$0.04$ variation across partitions. The centroid also changes across families and predictor settings, and mid-training checkpoints are about $0.05$ higher than converged ones. Panel~(b) asks whether the useful severity region changes with the final-training budget rather than remaining fixed. At each budget, the empirical severity centroid is computed over the probe-feasible regions, replacing $\crdgs$ by positive $U_{\mathrm{real}}$, using per-bin retraining at $0.5{\times}$, $1{\times}$, and $2{\times}$ the final budget. Within those regions, its centroid shifts by $+0.20$ for fog and $+0.19$ for low light from $0.5{\times}$ to $2{\times}$, while the scaled-$K$ $\crdgs$ prediction remains within $0.03$ of the empirical centroid at every budget.

\section{Discussion and Limitations}
\label{sec:discussion}

The twin-based comparisons isolate two different confounds.
Paired gap reduction is affected by clean forgetting, whereas $D_g$ still includes progress obtainable from clean training.
Subtracting the matched clean-control trajectory gives \crdg{}, which best tracks the signed retraining utility in our experiments.
The correctable band depends on the learner and the budget rather than on severity alone, and it shifts with the final budget as the scaled-$K$ score predicts.
The procedure transfers across the tested predictors unchanged, whereas the scores are recomputed for each model.

\textbf{Limitations.}
\crdg{} is an empirical surrogate for the full-training gain relative to clean-control training, tested at three budgets on one backbone.
Residual drift near the validity threshold may introduce mild label noise despite the structural gate.
Unseen families such as snow benefit only through cross-family transfer.
The probe horizon follows a fixed budget-scaling rule rather than adapting to each region.
The SOD twin pool comes from external clean images. Also, all newly trained models are single runs, and we do not report seed variance.

\section{Conclusion} 
\label{sec:conclusion}

We study how to curate synthetic degradations under a finite training budget.
\crdg{} compares degradation training with an equally long clean-control probe on held-out twins and treats clean harm separately.
Using this score, \method{} selects a fixed-budget training set and improves SOD and semantic segmentation across the tested predictors and generation settings.
We also find that the useful severity range changes with model state and training budget instead of remaining fixed.

\FloatBarrier
\bibliography{main}

\newpage
\appendix
\crefalias{section}{appendix}
\crefalias{subsection}{appendix}
\crefalias{subsubsection}{appendix}
\setcounter{figure}{0}
\renewcommand{\figurename}{Fig.}
\renewcommand{\thefigure}{S\arabic{figure}}
\setcounter{table}{0}
\renewcommand{\tablename}{Tab.}
\renewcommand{\thetable}{S\arabic{table}}

\section{Implementation Details}
\label{app:impl}

\textbf{Backbones and optimization.}
SemSeg follows the released ISSA configuration~\citep{li2023issa} for SegFormer. The DeepLabV3+ experiments follow the official RobustNet recipe~\citep{choi2021robustnet}. SOD clean-retention runs use NUN on a PVTv2-B2 backbone with its released training configuration. \method{} leaves the host optimizer, input resolution, augmentation, and final-training schedule unchanged in every experiment. All runs use a single NVIDIA RTX PRO 6000 GPU with 96\,GB of memory.

\textbf{Curation budget and probe settings.}
For the Cityscapes experiments, $B_{\mathrm{sel}}=5{,}950$ synthetic samples are filled in rank order over $32$ regions (four families, eight severity bins), with a $1{:}1$ clean/degraded slot ratio. Each region starts from $2{,}975\times2$ candidates before validity filtering. Probes optimize the first two terms of Eq.~\eqref{eq:trainloss} at the same slot ratio and $0.25{\times}$ the base learning rate, for $K$ equal to $1/80$ of the corresponding final-training schedule. Plug-in experiments keep each host's synthetic budget, and \crdg{} is recomputed per host. Under this budget, rank-greedy filling is concentrated. The top-ranked region contributes over $90\%$ of $\mathcal{S}^\star$ and the remainder comes from the next-ranked region. For WXSOD and CSOD10K the synthetic budget is filled entirely from the top-ranked feasible region, so $k$-center controls diversity within that region.

\textbf{Thresholds.}
$\lambda_s{=}1.0$, $\lambda_c{=}0.5$, $\lambda_b{=}1.0$ in Eq.~\eqref{eq:trainloss}. 
The clean-retention tolerance $\epsilon$ is set to one standard error of the clean-harm estimate from degradation-free matched probes on a held-out source split, where the standard error is taken over the per-image loss differences on the held-out images. The rule is fixed across experiments, and the value is estimated once for each task and predictor on its own source data.
We use $\kappa{=}1/2$ for the diagnostic band and $n_z{=}4$ regeneration draws in the validity fallback. Nothing is tuned on a target set.

\textbf{Evaluation protocol.}
SemSeg follows the ISSA evaluation protocol on Cityscapes val, the four ACDC conditions, and Dark Zurich. SOD benchmarks use their official metrics. Source-only and $M_{\mathrm{clean}}$ are obtained from the same $\theta_0$ with the same final-training schedule, with the synthetic slot replaced by clean data.
Unless otherwise noted, newly trained models are reported from single runs, and published reference values are taken from the cited sources. Operational curation uses one fixed probe seed and one fixed $A|B$ partition. Two additional $A|B$ partitions serve only the partition-stability analysis and the error bars of \Cref{fig:evidence}c and never enter curation (\Cref{app:probe}).

\section{Generator and Validity Gate}
\label{app:gate}

\textbf{Structure-preserving generation.}
We use Stable Diffusion 2.1 in image-to-image mode with depth- and Canny-conditioned ControlNet guidance together with the source image, but not the ground-truth mask, which avoids mask-aligned synthesis artifacts. We instantiate low light, backlight, fog, and rain. Each family contains eight ordered severity bins obtained by uniformly sampling its family-specific degradation-strength parameter, with two stochastic draws per source image. The strength variable of each family and its eight values are provided in the released generation configuration.

\textbf{Validity gate.}
The gate rejects candidates whose structure has drifted. Given $(x,\tilde{x},y)$ it measures frozen DINOv2 ViT-B/14 feature correspondence and boundary displacement in pixels within a band around ground-truth boundaries. For SOD it additionally checks that the salient object is neither deleted nor invented. It is a local structural test rather than a global appearance distance, so a uniformly darker or foggier image with intact geometry passes. Thresholds are set to the 99th percentile of the corresponding statistics on clean source pairs under geometric re-augmentation, after compensating for the known geometric transform, and reused across datasets. In a manual audit of $500$ candidates per family (\Cref{tab:gate}), acceptance is $98$--$100\%$ at mild levels and remains $90$--$96\%$ at severe levels, so the gate does not simply remove the stronger degradations from the pool.

\begin{algorithm}[htbp!]
	\caption{\method: controlled curation by reducibility}
	\label{alg:curb}
	\begin{algorithmic}[1]
		\Require clean source set $\mathcal{D}_s$, generator $\mathcal{G}$, checkpoint $\theta_0$, probe budget $K=\tfrac{1}{80}$ of the final schedule, tolerance $\epsilon$, sample budget $B_{\mathrm{sel}}$
		\State split source IDs once into $A|B$ and generate gated degradation
		regions $g=(d,s)$
		\State $\theta^{\mathrm{ctrl}}_{A}\gets U(\theta_0,A_{\mathrm{clean}}\oplus A_{\mathrm{aug}};K)$ and $\theta^{\mathrm{ctrl}}_{B}\gets U(\theta_0,B_{\mathrm{clean}}\oplus B_{\mathrm{aug}};K)$
		\Comment{shared controls}
		\ForAll{regions $g=(d,s)$}
		\State $\theta^{\mathrm{deg}}_{A,g}\gets U(\theta_0,A_{\mathrm{clean}}\oplus A_{\mathrm{deg},g};K)$ and likewise $\theta^{\mathrm{deg}}_{B,g}$
		\State compute $G^{A\to B}_g,H^{A\to B}_g$ on $B$ and $G^{B\to A}_g,H^{B\to A}_g$ on $A$
		\State $\crdgs(g;K,\theta_0)\gets \tfrac12(G^{A\to B}_g+G^{B\to A}_g)$
		\State $\mathrm{feasible}(g)\gets [\max(H^{A\to B}_g,H^{B\to A}_g)\le\epsilon]$
		\EndFor
		\State rank feasible regions with positive $\crdgs(g;K,\theta_0)$ and fill $B_{\mathrm{sel}}$ in descending order, exhausting each region before moving to the next
		\State use $k$-center when only part of a region is needed
		\State use clean data for any unfilled budget and obtain $\mathcal{S}^{\star}$
		\State train from $\theta_0$ on clean data $\cup\,\mathcal{S}^{\star}$ with Eq.~\eqref{eq:trainloss}
	\end{algorithmic}
\end{algorithm}

\begin{table}[htbp]
	\centering
	\footnotesize
	\setlength{\tabcolsep}{5pt}
	\begin{tabular}{@{}lcccc@{}}
		\toprule
		Degradation & gate prec. & gate recall & \makecell{accept.\\$s_1$--$s_4$} & \makecell{accept.\\$s_5$--$s_8$} \\
		\midrule
		Low light & 0.94 & 0.91 & 0.99 & 0.95 \\
		Backlight & 0.92 & 0.89 & 0.99 & 0.92 \\
		Fog       & 0.95 & 0.92 & 1.00 & 0.96 \\
		Rain      & 0.91 & 0.88 & 0.98 & 0.90 \\
		\bottomrule
	\end{tabular}
	\caption{\textbf{Validity-gate audit} on $500$ sampled candidates per family. Structural drift is the positive class for gate precision and recall. Acceptance remains high at severe levels ($0.90$--$0.96$), indicating that the gate does not simply remove stronger degradations.}
	% \vspace{-3mm}
	\label{tab:gate}
\end{table}

\section{Probe Protocol and Estimator Details}
\label{app:probe}

\textbf{Probe matching.}
Operational curation uses one fixed random split of source image IDs into equal halves $A$ and $B$, shared across all regions. For each region $g$ and split direction, let $S\in\{A,B\}$ denote the probe-update half. The treatment branch trains on $S_{\mathrm{clean}}\oplus S_{\mathrm{deg},g}$ and the control branch on $S_{\mathrm{clean}}\oplus S_{\mathrm{aug}}$, both from $\theta_0$. The two branches share $K$, batch composition, clean-replay ratio, optimizer and schedule, geometric views, and data order. The region-independent control is run once per direction and shared across regions. One fixed probe seed is used operationally. Two additional $A|B$ partitions serve only to assess stability and provide the error bars of \Cref{fig:evidence}c, and probe horizons of $0.5\times$ and $2\times$ the default serve only the sensitivity analysis. \Cref{alg:curb} gives the full procedure. The \emph{training-twin} variant scores a region by the loss drop on the treatment probe's own degraded training twins rather than by held-out transfer. The \emph{in-sample} variant keeps the controlled gain $G_g$ but evaluates both branches on the probe-update half instead of the held-out half.

\textbf{Matched views and validity fallback.}
Treatment and control views share the crop and flip of each source image. 
The control applies weak photometric augmentation (colour jitter, gamma in $[0.8,1.25]$, and Gaussian blur with $\sigma\le0.6$), while the treatment uses a generated twin from region $g$.
Twins that fail the gate are resampled up to $n_z{=}4$ times. If no valid training twin is found, the control view is retained in that slot, and invalid evaluation twins are dropped from both branches.

\textbf{Statistical reporting.}
$\crdgs$ is reported signed rather than rectified, preserving negative transfer. Aggregate statistics use paired bootstrap over regions. The bootstrap of \Cref{fig:evidence}b resamples ACDC evaluation images with replacement within each condition, using matched indices for every retrained model and the clean model. Checkpoints are kept fixed, and $U_{\mathrm{real}}$, band membership, and the centroid are recomputed per replicate. Error bars in \Cref{fig:evidence}c are standard deviations over the three partitions.

\textbf{Sensitivity to $\epsilon$.}
\Cref{tab:eps} varies the clean-retention tolerance around its default of one standard error. Degraded-domain performance is stable over $0.5$--$2$ standard errors, while a looser tolerance admits more regions at the cost of a small drop in clean accuracy on Cityscapes val.

\begin{table}[htbp]
	\centering
	\footnotesize
	\setlength{\tabcolsep}{5pt}
	\begin{tabular}{@{}lcccc@{}}
		\toprule
		$\epsilon$ & $0$ & $0.5\,\mathrm{SE}$ & $1\,\mathrm{SE}$ (default) & $2\,\mathrm{SE}$ \\
		\midrule
		feasible-region fraction & 0.41 & 0.63 & 0.78 & 0.91 \\
		in-family mIoU $\uparrow$ & 51.43 & 51.86 & \textbf{51.97} & 51.79 \\
		Cityscapes mIoU $\uparrow$ & 70.21 & 70.24 & 70.13 & 69.82 \\
		\bottomrule
	\end{tabular}
	\caption{\textbf{Sensitivity to the clean-retention tolerance $\epsilon$.} Degraded-domain performance is stable around the default, while loosening the constraint permits a small clean-performance drop.}
	\label{tab:eps}
\end{table}

\section{Plug-and-Play Framework Comparisons}
\label{app:plugin}

\textbf{Protocol.}
\Cref{tab:plugin} follows each host's published pipeline and synthetic-data budget. 
Published references for ISSA, H-Weather, and RobustNet+ISSA are taken from \citet{li2023issa}. Gen4Seg follows \citet{yin2026gen4seg}, with its Dark Zurich value obtained using the released pipeline. For our ``+\method{}'' runs, \method{} is applied under the corresponding host setting and synthetic-data budget, with \crdg{} recomputed for each host.

\textbf{Region construction.}
Regions are defined by family and severity bin, with eight bins per family unless the host provides fewer discrete levels. ISSA~\citep{li2023issa} uses style-exemplar groups and style-mixing strength. Hendrycks-Weather~\citep{hendrycks2019benchmarking} uses corruption type and its predefined levels $1$--$5$. Gen4Seg~\citep{yin2026gen4seg} uses weather type and a family-specific image-space statistic relative to the source (contrast attenuation for fog, luminance ratio for night, high-frequency energy gain for rain, bright-region coverage for snow), with prompt strength not used for region assignment. RobustNet + ISSA uses the ISSA regions with \crdg{} recomputed on the RobustNet model.

\section{SOD Benchmark Protocols}
\label{app:sodprotocol}

WXSOD~\citep{wxsod2025} and CSOD10K~\citep{cong2025breaking} use their official splits and metrics, with each base model following its published training recipe. The augmented variants add twins synthesized from clean DUTS-TR with the four families and eight severity levels of \Cref{app:gate}, including its validity gate. The uncurated and \method{} variants use the same synthetic budget and the base model's final-training schedule. The former samples uniformly, and the latter selects with \crdg{}. We use $B_{\mathrm{sel}}=4{,}000$ on WXSOD and $3{,}750$ on CSOD10K.

\section{Compute and Cost Accounting}
\label{app:cost}

For the reported SegFormer configuration, the $66$ probe updates cost approximately $0.59\times$ one final-training run under the $K=T/80$ scaling rule. Candidate-pool construction, held-out scoring, and selection are not included, and final training is accounted separately in \Cref{sec:exp_cost}.

\begin{table}[htbp]
	\centering
	\footnotesize
	\setlength{\tabcolsep}{6pt}
	\begin{tabular}{@{}lr@{}}
		\toprule
		Quantity & Value \\
		\midrule
		Regions $M$ (four families $\times$ eight severity bins) & $32$ \\
		Split directions $R$ & $2$ \\
		Probe horizon $K$ & $1/80$ of the final schedule \\
		Treatment probes ($MR$) & $64$ \\
		Shared control probes ($R$) & $2$ \\
		Probe-step cost / final-step cost (measured) & $\approx0.71$ \\
		Probe-update cost / one final run & $0.59\times$ \\
		Control overhead vs.\ uncontrolled probes ($2/64$) & $3.1\%$ \\
		\bottomrule
	\end{tabular}
	\caption{\textbf{Curation cost.} Operational selection uses one $A|B$ partition. The region-independent control adds only two probes, and the normalized probe-update cost is $(66/80)\times0.71\approx0.59$ of one final-training run, under the $K=T/80$ scaling rule.}
	\label{tab:cost}
\end{table}

\section{Baseline Adaptation}
\label{app:full}

All selection baselines rank the same $32$ gated regions and use the same $B_{\mathrm{sel}}$ and within-region $k$-center step. Candidate-level scores are averaged within each region before ranking, and the final training recipe is identical for every method. RHO-Loss-style selection uses the reducible loss of a candidate relative to a small model trained on held-out source data. Gradient contribution uses the alignment of a candidate's gradient with the mean source-validation gradient. The DisCL-style baseline is converted to a static region score. Its curriculum rule is run on the source-validation response of the current model to obtain a severity preference per family, which is then used to rank regions once before final training rather than to alter the training schedule.

\end{document}